# Current Time Series Anomaly Detection Benchmarks are Flawed and are Creating the Illusion of Progress

Renjie Wu and Eamonn J. Keogh

**Abstract**—Time series anomaly detection has been a perennially important topic in data science, with papers dating back to the 1950s. However, in recent years there has been an explosion of interest in this topic, much of it driven by the success of deep learning in other domains and for other time series tasks. Most of these papers test on one or more of a handful of popular benchmark datasets, created by Yahoo, Numenta, NASA, etc. In this work we make a surprising claim. The majority of the individual exemplars in these datasets suffer from one or more of four flaws. Because of these four flaws, we believe that many published comparisons of anomaly detection algorithms may be unreliable, and more importantly, much of the apparent progress in recent years may be illusionary. In addition to demonstrating these claims, with this paper we introduce the UCR Time Series Anomaly Archive. We believe that this resource will perform a similar role as the UCR Time Series Classification Archive, by providing the community with a benchmark that allows meaningful comparisons between approaches and a meaningful gauge of overall progress.

## 1 Introduction

TIME series anomaly detection has been a perennially important topic in data science, with papers dating back to the dawn of computer science [1]. However, in the last five years there has been an explosion of interest in this topic, with at least one or two papers on the topic appearing each year in virtually every database, data mining and machine learning conference, including SIGKDD [2], [3], ICDM [4], ICDE, SIGMOD, VLDB, etc.

A large fraction of this increase in interest seems to be largely driven by researchers anxious to transfer the considerable success of deep learning in other domains and from other time series tasks such as classification.

Most of these papers test on one or more of a handful of popular benchmark datasets, created by Yahoo [5], Numenta [6], NASA [2] or Pei's Lab (OMNI) [3], etc. In this work we make a surprising claim. The majority of the individual exemplars in these datasets suffer from one or more of four flaws. These flaws are *triviality, unrealistic anomaly density, mislabeled ground truth* and *run-to-failure bias*. Because of these four flaws, we believe that most published comparisons of anomaly detection algorithms may be unreliable. More importantly, we believe that much of the apparent progress in recent years may be illusionary.

For example, Qiu *et al.* [7] introduce a "*novel anomaly detector for time-series KPIs based on supervised deep-learning models with convolution and long short-term memory (LSTM) neural networks, and a variational auto-encoder (VAE) oversampling model.*" This description sounds like it has many "moving parts", and indeed, the dozen or so explicitly listed parameters include: convolution filter, activation, kernel size, strides, padding, LSTM input size, dense input size, softmax loss function, window size, learning rate and batch size. All of this is to demonstrate "*accuracy exceeding 0.90 (on a subset of the Yahoo's anomaly detection benchmark datasets).*" However, as we will show, much of the results of this complex approach can be duplicated with a single line of code and a few minutes of effort.

This "one-line-of-code" argument is so unusual that it is worth previewing it before we formally demonstrate it in Section 2.2 below. Almost daily, the popular press vaunts a new achievement of deep learning. Picking one at random, in a recent paper [8], we learn that deep learning can be used to classify mosquitos' species. In particular, the proposed algorithm had an accuracy of 97.8% when distinguishing *Aedes vexans* from *Culex triaeniorhynchus*. Should we be impressed? One of the current authors (Keogh) has significant computational experience working with mosquitos, and he *is* impressed.

Suppose however that someone downloaded the original 1,185 images from the study and showed that they could classify them with 100% accuracy using a single line of code[1]. If that happened, there are two things we can confidently say:

- We would not for one moment imagine that the one line of code had any particular value as a classifier. We would assume that this was some kind of "trick". Perhaps the *Aedes* images were in JPEG

- *R. Wu is with the Department of Computer Science and Engineering, University of California Riverside, Riverside, CA 92521. E-mail: rwu034@ucr.edu.*
- *E.J. Keogh is with the Department of Computer Science and Engineering, University of California Riverside, Riverside, CA 92521. E-mail: eamonn@cs.ucr.edu.*

[1] To be clear, we choose this example because it was the first hit for a Google search for "novel deep learning applications". We have no reason to doubt the claims of this paper, which we only skimmed.

format and the *Culex* images were in GIF format. Or perhaps one species was recorded in color, and the other in B/W. Something interesting is clearly happening, but it is surely not the case that a useful entomological image classification algorithm takes a single line of code.

- We would have lost some confidence in the original paper's results. It is still likely that the paper is genuinely doing something useful. However, we would all be a lot happier trusting the paper's contribution if the authors released a statement to the effect of "*we converted all files to JPEG format, and all images to 16-bit B/W, and reran the experiments getting similarly good results. Moreover, we are confident that our new publicly released dataset will now not yield to a single line of code algorithm*".

This is a perfect analogy of our one-line-of-code argument. Our ability to produce "one-liners" for most datasets does not mean that the original papers that tested on these datasets are not making a contribution. However, at a minimum it does *strongly* suggest that the community needs to regroup, and test on new datasets that would generally stump trivial one-line solutions.

Before continuing it is important to note that our discussion of some issues with the benchmark datasets should in no way be interpreted as criticism of the original introducer of these datasets. These groups have spent tremendous time and effort to make a resource available to the entire community and should rightly be commended. It is simply the case that the community must be aware of the severe limitations of these datasets, and the limitations of any research efforts that rely upon them.

## 2 A Taxonomy of Benchmark Flaws

Before discussing the four major flaws found in many public archives, we will briefly discuss related work, to put our observations into context.

### 2.1 Related Work

The literature on anomaly detection is vast [9], [10] with a particular increase in works in just the last three to five years [2], [4], [5], [6], [7], [11], [12], [13], [14], [15]. We refer the interested reader to [10] which offers the reader a detailed review and taxonomy.

Almost all these works test on one or more public datasets created by a handful of groups, including Yahoo [5], Numenta [6], NASA [2] or OMNI [3]. Some papers test on these public datasets in addition to a private dataset. In many cases, the authors do not even show a plot of any data from the private datasets. Thus, here we can clearly make no claims about such private datasets, other than to note that the use of private datasets thwarts the community's laudable move to reproducibility. In addition, the use of private datasets will always be accompanied by the possibility of unconscious cherry-picking that the reader or the reviewer will never know about.

There is a strong implicit assumption that doing well on one of the public datasets is a sufficient condition to declare an anomaly detection algorithm useful (and therefore warrant publication or patenting). Indeed, this assumption is stated explicitly in many works, for example Huang [16] notes "(The Yahoo) *A1Benchmark is undoubtedly a good time-series dataset for testing the general effectiveness of an anomaly detection method*", and Gao *et al.* [17] gush that "*Yahoo data set has a good coverage of different varieties of anomalies in time series, such as seasonality, level change, variance change and their combinations*." However, we are not aware of any work that has treated this assumption critically.

In the following four sections, we will introduce four issues with these public datasets that we believe throws doubt on the assumption that they are suitable for comparing algorithms or gauging progress in time series anomaly detection.

### 2.2 Triviality

A large fraction of the problems in the benchmark datasets are so simple to solve that reporting success in solving them seems pointless or even absurd. Of course, *trivial* is not a well-defined word, so, to firm up our claim we will make a practical testable definition:

**Definition 1.** A time series anomaly detection problem is *trivial* if it can be solved with a single line of standard library MATLAB code. We cannot "cheat" by calling a high-level built-in function such as *kmeans* or *ClassificationKNN* or calling custom written functions. We must limit ourselves to basic vectorized primitive operations, such as *mean*, *max*, *std*, *diff*, etc.

This definition is clearly not perfect. MATLAB allows nested expressions, and thus we can create a "one-liner" that might be more elegantly written as two or three lines. Moreover, we can use unexplained "magic numbers" in the code, that we would presumably have to learn from training data. Finally, the point of anomaly detectors is to produce purely automatic algorithms to solve a problem. However, the "one-liner" challenge requires some human creativity (although most of our examples took only a few seconds and did not tax our ingenuity in the slightest).

Nevertheless, our simple definition gets at the heart of our point. If we can quickly create a simple expression to separate out anomalies, it strongly suggests that it was not necessary to use several thousands of lines of code and tune up to a dozen parameters to do it.

Perhaps the best way to see this is to imagine that we give the same challenge to create a "one-liner" for differentiating protein-coding and noncoding RNA [14], or we had the challenge of separating positive vs negative Yelp reviews. Both of these are also one-dimensional problems on which deep learning appears to have made significant progress in recent years [14]. However, it seems inconceivable that the bioinformatic or text datasets considered in the literature could be teased apart with a single line of code, no matter how contrived. These are intrinsically hard problems, and the communities working on them are using intrinsically challenging datasets.

To illustrate our point, consider Fig. 1, which shows an example from the OMNI dataset [3]. The example is a multiple-dimensional dataset, here we consider only dimension 19.

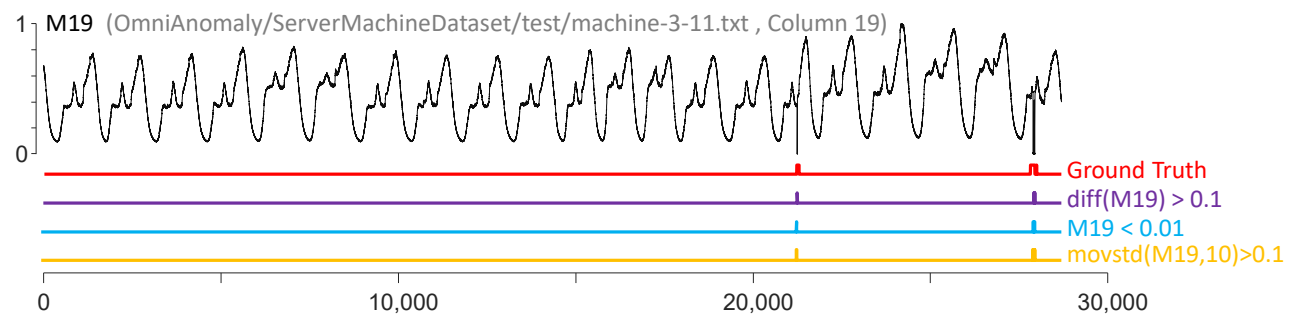

Fig. 1. *(top to bottom)* Dimension 19 from SDM3-11 dataset. A binary vector (red) showing the ground truth anomaly labels. Three examples of "one-liners" that can solve this problem.

There are dozens of simple one-liners that solve this problem. In the figure we show three representative examples.

Let us take the time to preempt some possible objections to this demonstration.

- *All the one-liners have a parameter.* True, but recall that most anomaly detection algorithms, especially ones based on deep learning, have *ten* or more parameters. Moreover, the results here are not particularly sensitive to the parameter we set.
- *The choice of dimension was cherry-picked*. We deliberately chose one of the *harder* of the 38 dimensions here. Most of the rest are even *easier* to solve.
- *The choice of problem was cherry-picked.* Of the twenty-eight example problems in this data archive, the majority are this easy to solve with one-liners.
- *The fact that you can solve this problem in one line, does not mean that other algorithms that are successful in this dataset are not useful.* True, we have acknowledged that point in multiple places in this work and are happy to do so again here.

The second most cited benchmark is Numenta [6]. The Numenta archive is commendably diverse, however most of the examples, like the one shown in Fig. 2, readily yield to a single line of code.

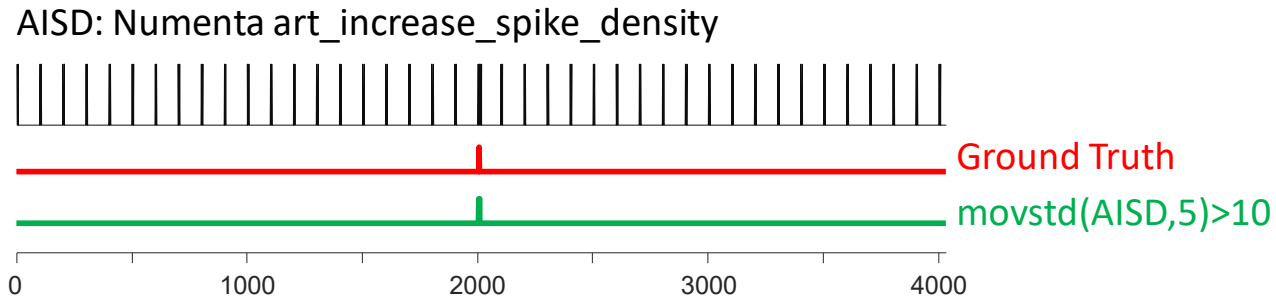

Fig. 2. *(top to bottom)* The Numenta Art Increase Spike Density datasets. A binary vector (red) showing the ground truth anomaly labels. A "one-liner" (green) that can solve this problem.

We will not even bother to show any examples from the NASA dataset (the interested reader can view many examples on [18]). In about half the cases the anomaly is manifest in many orders of magnitude difference in the value of the time series. Such examples are well beyond trivial.

Other NASA examples consist of a dynamic time series suddenly becoming exactly constant (see in Fig. 9). For those examples, we can flag an anomaly if, say, three consecutive values are the same, with something such as `diff(diff(TS)) == 0`.

Having said that, perhaps 10% of the examples in the NASA archive are mildly challenging, although even those examples do not need to avail of the power of deep learning, as the yield to decade-old simple ideas [19].

The Yahoo archive [5] is by far the most cited in the literature. It contains a mixture of real and synthetic datasets. Let us consider the first real dataset, which happens to be one of the more challenging examples (at least to the human eye). However, as Fig. 3 shows, it readily yields to a one-liner.

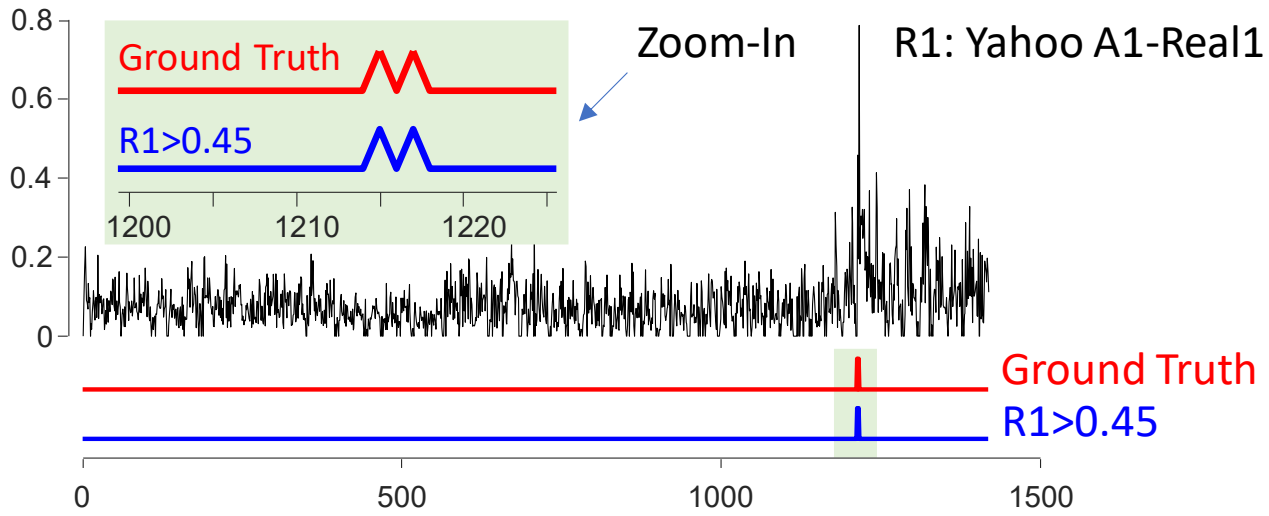

Fig. 3. Yahoo A1-Real1. A binary vector (red) showing the ground truth anomaly labels. An example of a "one-liner" (blue) that can solve this problem. A zoom-in shows how precisely the simple one-linear can match the ground truth.

Lest the reader think that we cherry-picked here, let us consider the *entire* Yahoo Benchmark [5]. There are 367 time series in the Yahoo Benchmark; most of them can be solved with a *universal* one-liner (1) or (2):

```
abs(diff(TS)) > u * movmean(abs(diff(TS)), k)
              + c * movstd(abs(diff(TS)), k)
              + b                                   (1)
```

```
diff(TS) > u * movmean(diff(TS), k)
         + c * movstd(diff(TS), k)
         + b                                        (2)
```

where **TS** is the time series, **u** is either 0 or 1 to determine whether `movmean` is used, **k** is the window size to compute **k**-points mean values and standard deviations, **c** is the coefficient applied to `movstd`, and **b** is the offset to adjust the center of the right-hand side of (1) or (2).

The only difference between (1) and (2) is to use either `diff(TS)` or `abs(diff(TS))`. From (1) and (2), we can derive the following simplified one-liners:

```
abs(diff(TS)) > b                                   (3)
abs(diff(TS)) > movmean(abs(diff(TS)), k)
              + c * movstd(abs(diff(TS)), k)
              + b                                   (4)
```

```
diff(TS) > b                                        (5)
diff(TS) > movmean(diff(TS), k)
         + c * movstd(diff(TS), k)
         + b                                        (6)
```

We did a simple bruteforce search to compute individual **k**, **c** and **b** which solve anomaly detection problems on all 367 time series. As the results show in Table 1, we are surprised by the triviality of the Yahoo Benchmark: 316 out of 367 (86.1%) can be easily solved with a one-liner.

TABLE 1
Bruteforce results on Yahoo Benchmark

| Dataset | Solvable with | # Time Series Solved | # Time Series in Dataset | Percent |
|---|---|---|---|---|
| A1 | (3) | 30 | 67 | 44.8% |
| | (4) | 14 | | 20.9% |
| | Subtotal | 44 | 67 | 65.7% |
| A2 | (3) | 40 | 100 | 40.0% |
| | (4) | 57 | | 57.0% |
| | Subtotal | 97 | 100 | 97.0% |
| A3 | (5) | 84 | 100 | 84.0% |
| | (6) | 14 | | 14.0% |
| | Subtotal | 98 | 100 | 98.0% |
| A4 | (5) | 39 | 100 | 39.0% |
| | (6) | 38 | | 38.0% |
| | Subtotal | 77 | 100 | 77.0% |
| | Total | 316 | 367 | 86.1% |

Surprisingly, 193 out of 367, that is more than half, time series in Yahoo Benchmark can be solved with individual *magic numbers* **b** in (3) or (5). Even for those fourteen time series solvable with (6) in A3 dataset, they share a common property of **k** = 5 and **c** = 0, while **b** varies case by case.

The overall 86.1% number seems competitive with most papers that have examined this dataset [7], [16], [17] (it is difficult to be more precise than that because of the vagaries of scoring functions). Moreover, as we will show in Section 2.4, because of some labeling errors, this is probably as close to perfect as can be achieved on this dataset.

In [18] we show a gallery of dozens of additional examples from Yahoo [5], Numenta [6], NASA [2] and Pei's Lab (OMNI) [3] that yield to one line solutions.

## 2.3 Unrealistic Anomaly Density

This issue comes in three flavors:

- For some examples, more than half the test data exemplars consist of a contiguous region marked as anomalies. For example, NASA datasets D-2, M-1 and M-2. Another dozen or so have at least 1/3 of their length consist of a contiguous region marked as anomalies [2].
- For some examples, there are *many* regions marked as anomalies. For example, SDM exemplar machine-2-5 has 21 separate anomalies marked in a short region.
- In some datasets, the annotated anomalies are very close to each other. For example, consider Fig. 3, it shows two anomalies sandwiching a *single* normal datapoint.

There are many issues with such an unrealistic anomaly density. First, it seems to blur the line between *classification* and *anomaly detection*. In most real-world settings, the prior probability of an anomaly is expected to be only slightly greater than zero. Having half the data consist of anomalies seems to violate the most fundamental assumption of the task. Moreover, many algorithms are very sensitive to the priors.

Another issue is that this unrealistic density greatly confuses the task of scoring and comparing algorithms. Suppose we have a dataset with ten anomalies, one at about midnight for ten days, reflecting an increasingly weakening pump filling a tank at the start of a batch process. We could imagine two rival algorithms, each of which managed to detect a single anomaly. However, one algorithm finds the first anomaly, and the other algorithm finds the last. These outcomes correspond to very different practical results when deployed. The former saves ten bad batches being created, the latter only one. We might imagine rewarding more for earlier detection, and in fact the Numenta team [6] (among others) have suggested that. However, the resulting scoring function is exceedingly difficult to interpret, and almost no one uses this [20].

We believe that the ideal number of anomalies in a single testing time series is exactly *one*. Moreover, this number should be communicated with the dataset. This makes the users task a little easier. Instead of trying to predict *if* there is an anomaly in the dataset, the algorithm should just return the most likely *location* of the anomaly. However, for this slight simplification which ignore specificity (which can and should be evaluated separately), we gain the fact that the evaluation is now *binary*. By testing on multiple datasets, we can report the aggregate results as simple *accuracy*, which is intuitively interpretable.

## 2.4 Mislabeled Ground Truth

All of the benchmark datasets appear to have mislabeled data, both false positives and false negatives. Of course, it seems presumptuous of us to make that claim, as the original creators of the datasets may have had access to out-of-band data they used to produce the labels. Nevertheless, we believe that many of the examples we claim are compelling enough to be unambiguous.

For example, consider the snippet of Yahoo A1-Real32 shown in Fig. 4. Any algorithm that points to location **B** will be penalized as having a false positive, but a true positive region **A**, is part of the same constant line. Since literally nothing has changed from **A** to **B**, it is hard to see how this labeling makes sense[2].

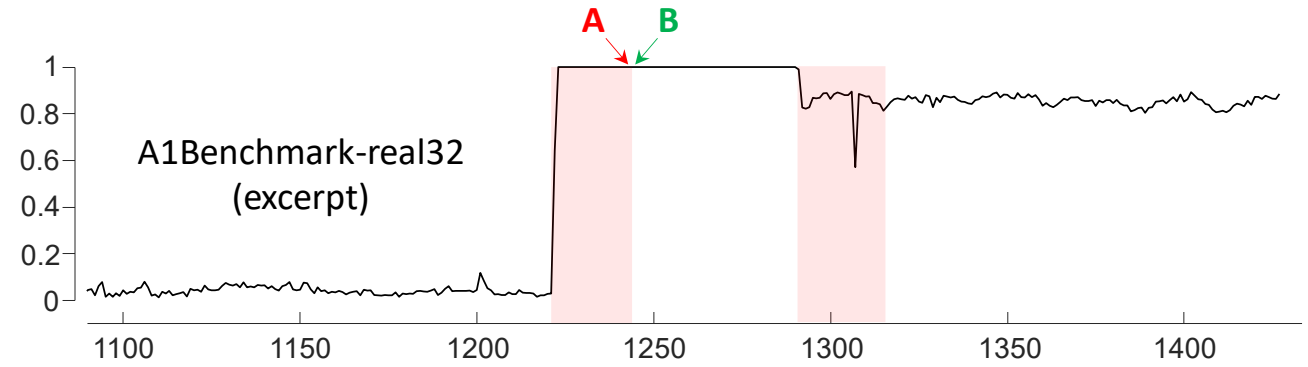

Fig. 4. An excerpt from Yahoo A1-Real32. An algorithm that points to **A** will be marked as a true positive. An algorithm that points to **B** will be marked as a false positive.

In Fig. 5 we see another Yahoo time series. There is a point anomaly (or "dropout") marked with **C**. However, at location 360 there is an almost identical dropout **D** that is not labeled as having an anomaly.

[2] If the rest of the data had many short constant regions, say of length 12, then you could imagine that a good algorithm might consider the 13th constant datapoint in a row an anomaly. However, this is the only constant region in this dataset.

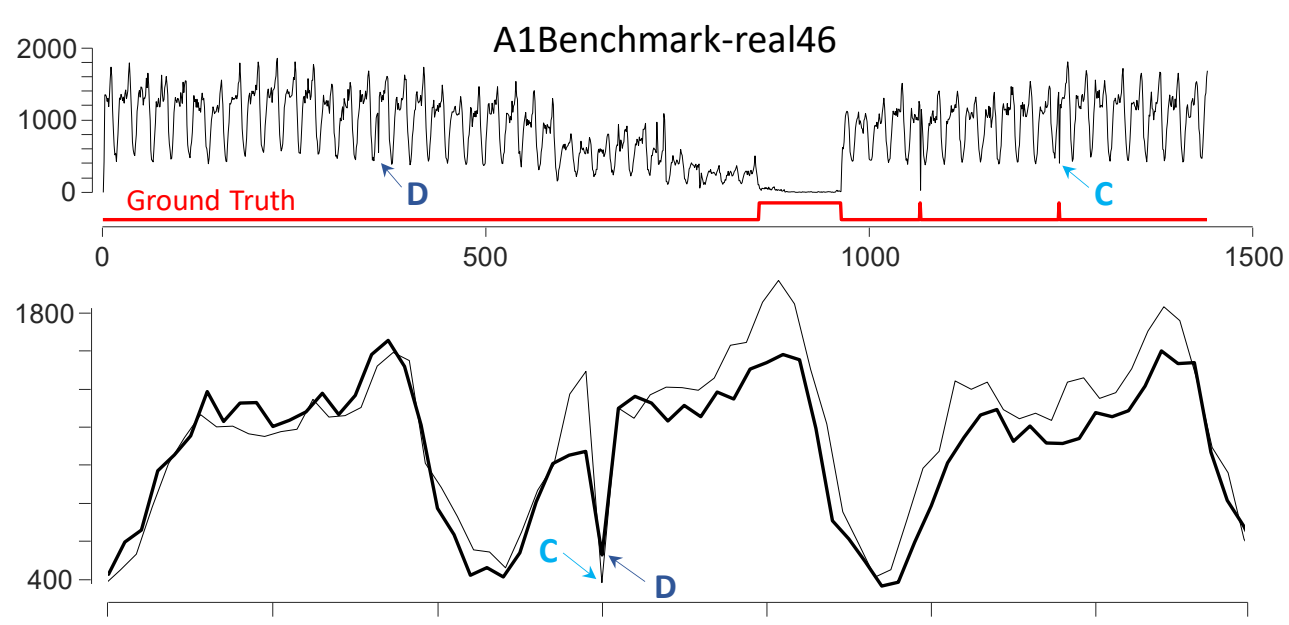

Fig. 5. *(top)* The Yahoo A1-Real46 dataset with its class labels (red). *(bottom)* Overlaying two snippets allows a one-to-one comparison between the region of **C** and **D**. The single point marked **C** is a true positive, but surprisingly, the point marked **D** is not.

In Fig. 6 we see a snippet of Yahoo A1-Real47 with two labeled anomalies. The one pointed to by **E** seems like a dropout, but **F** is a puzzle. Its rounded bottom visually looks like a dozen other regions in this example. If we measure its mean, min, max, variance, autocorrelation, complexity, Euclidean distance to the nearest neighbor, etc. and compare these numbers to other rounded bottom regions (Fig. 6 shows two others, of the about 48), there is simply nothing remarkable about it.

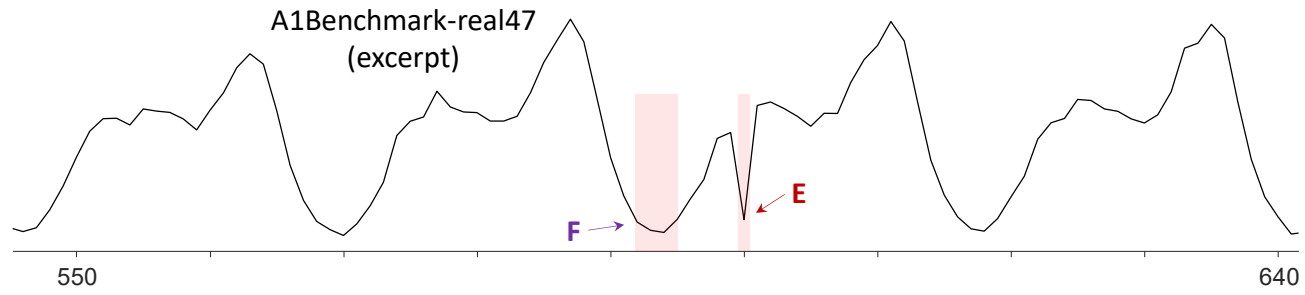

Fig. 6. An excerpt from Yahoo A1-Real47. Both **E** and **F** are marked as anomalies, but it is hard to see that **F** is truly an anomaly.

Beyond these issues, there are other labeling issues in the Yahoo datasets. For example, two datasets seem to be essentially duplicates (A1Real13 and A1Real15). An additional issue is more subjective, but some of the datasets seem to have unreasonably precise labels. Consider the labels shown in Fig. 7 *(top)*.

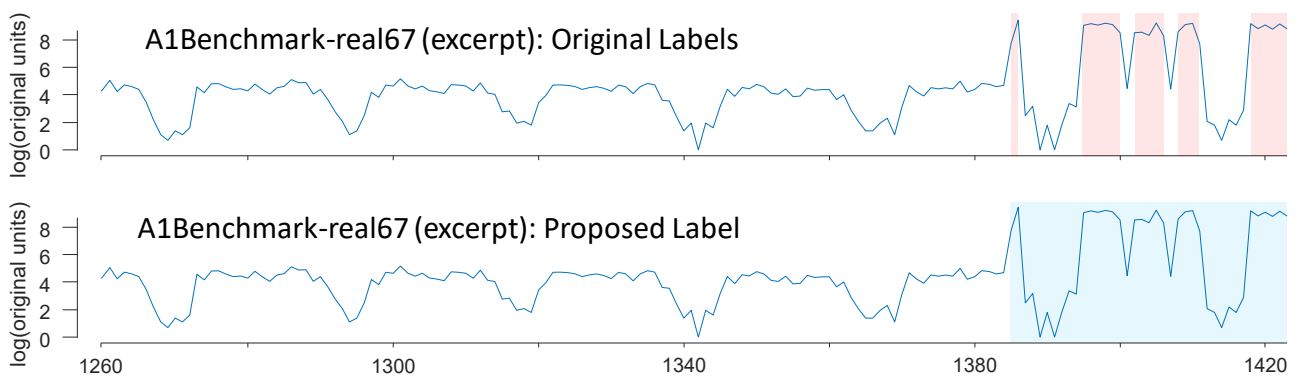

Fig. 7. *(top)* An excerpt from the Yahoo A1-Real67 dataset with its class labels (red). *(bottom)* Our proposed label (blue) for this dataset.

By analogy, some modern automobiles have anomaly detection sensors to detect violent crashes. Imagine a fast-moving car is involved in a crash and goes thumbing end-over-end down the highway. At some points in the rotation, the car will momentarily have a normal orientation. However, it would be bizarre to label those regions as "`normal`". Similarly, in A1-Real67 after about 50 almost identically repeated cycles, at time 1,384 the system has clearly dramatically changed, warranting flagging an anomaly. However, the subsequent rapid toggling between "`anomaly`" and "`normal`" seems unreasonably precise.

There are several reasons why this matters. Most anomaly detectors effectively work by computing statistics for each subsequence of some length. However, they may place their computed label at the beginning, the end or the middle of the subsequence. If care is not taken, an algorithm may be penalized because it reports a positive *just* to the left (or *just* to the right) of a labeled region. This is always a possible concern, but it becomes much more of an issue with rapid toggling of states.

One of the most referenced datasets is Numenta's NY Taxi data, which records the taxi demand in NY City from 2014/07/01 to 2015/01/31 [6]. According to the original labels, there are five anomalies, corresponding to the NYC marathon, Thanksgiving, Christmas, New Year's Day, and a blizzard. However, as shown in Fig. 8 this ground truth labeling seems to have issues.

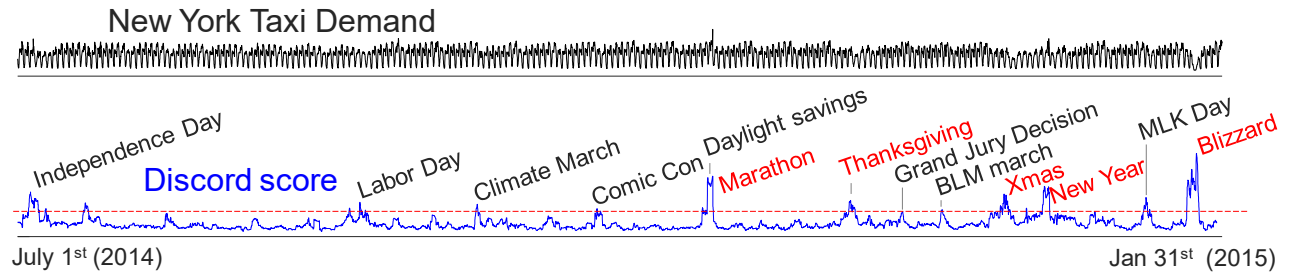

Fig. 8. *(top)* Numenta's NY Taxi dataset. *(bottom)* The *time series discord* score of the dataset [19], [21], with peaks annotated. The red text denotes the ground truth labels.

One minor issue is the anomaly attributed to the NYC marathon is really caused by a daylight-saving time adjustment that was made the same day.

However, the main problem with the five labels is that they seem very subjective. After a careful visual analysis, we believe that there are at least seven more events that are equally worthy of being labeled anomalies, including Independence Day, Labor Day and MLK Day. In addition to these USA holidays, we can easily detect the impromptu protests that followed the grand jury decision not to indict officers involved in the death of Eric Garner, "*Large groups shouted and carried signs through Times Square… Protesters temporarily blocked traffic in the Lincoln Tunnel and on the Brooklyn Bridge*" [22], and the more formal protest march that followed ten days later.

It is difficult to overstate the implications of this finding. At least dozens of papers have compared multiple algorithms on this dataset [6], [11], [12]. However, it is possible that an algorithm that was reported as performing very poorly, finding zero true positives and multiple false positives, actually performed *very* well, discovering Grand Jury, BLM march, Comic Con, Labor Day and Climate March, etc.

Finally, let us consider an example from the NASA archive [2]. In Fig. 9 we show three snippets from a test set. One of the snippets is labeled with the only anomaly acknowledged in this dataset. The anomaly corresponds to a dynamic behavior, becoming "frozen" for a period of time. However, the two other snippets also have this strange neighbor, but are *not* marked as anomalies. As always, it is possible that the creators of this archive have access to some out-of-band information that justifies this (none of the metadata or reports that accompany the data discuss this). However, in this case, it is particularly hard to believe these labels. In any case, suppose we compare two algorithms on this dataset. Imagine that one finds just the first true anomaly, and the other finds all three events highlighted in Fig. 9. Should we really report the former algorithm as being vastly superior?

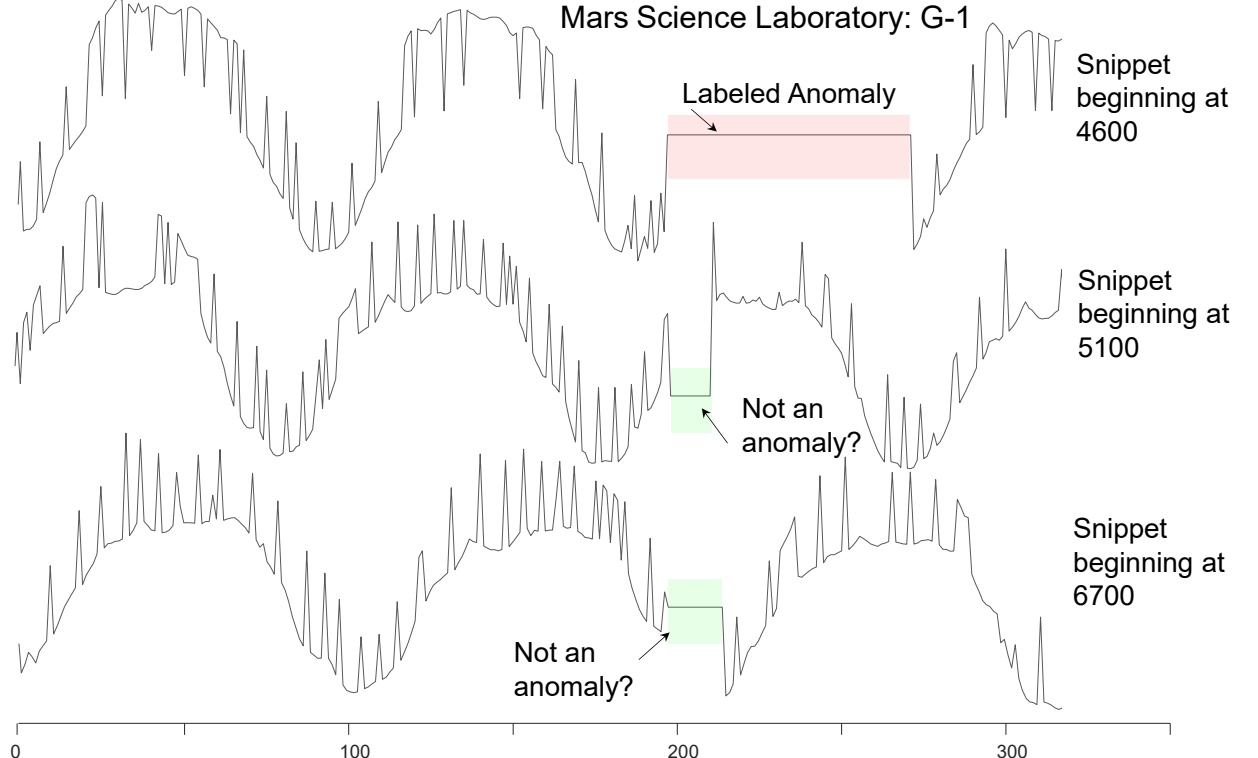

Fig. 9. (*top to bottom*) Three snippets from Mars Science Laboratory: G-1. The topmost one has the only labeled anomaly in this dataset. However, the bottom two snippets have essentially identical behaviors as the anomaly, but are not identified as such.

### 2.5 Run-to-failure Bias

There is an additional issue with at least the Yahoo (and NASA) datasets. As shown in Fig. 10, many of the anomalies appear towards the end of the test datasets.

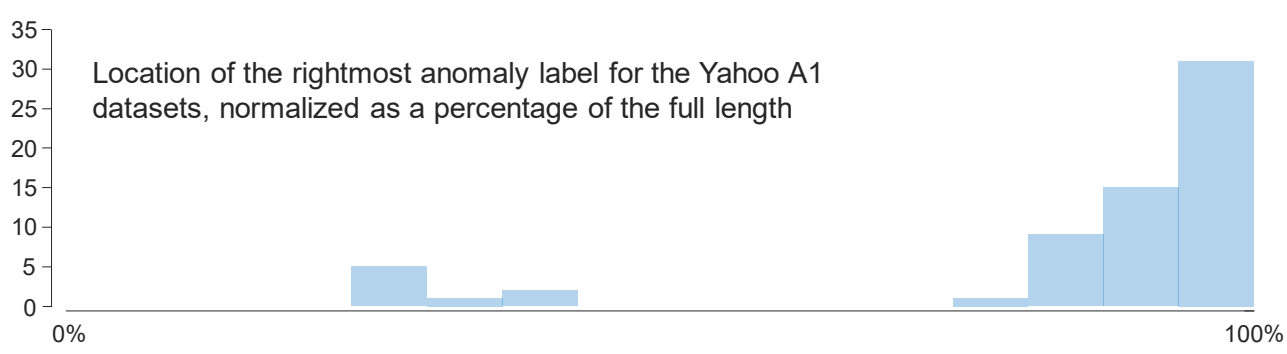

Fig. 10. The locations of the Yahoo A1 anomalies (rightmost, if there are more than one) are clearly not randomly distributed.

It is easy to see why this could be true. Many real-world systems are run-to-failure, so in many cases, there *is* no data to the right of the last anomaly. However, it is also easy to see why this could be a problem, as it drastically affects the default rate. A naïve algorithm that simply labels the last point as an anomaly has an excellent chance of being correct.

### 2.6 Summary of Benchmark Flaws

We believe that we have demonstrated that the classic time series anomaly detection archives are irretrievably flawed. For example, if we were told that algorithm *A* could achieve an F1 score of 1.0 on one of these datasets, should we be impressed? Given what we know about the amount of mislabeling on these datasets, we should not be impressed, instead we should have to suspect fraud or (much more plausibly) error.

However, suppose instead that we were told that algorithm *B* could achieve an F1 score of 0.9 on one of these datasets. Given what we know about the triviality of these datasets, this seems like something we could match or beat with decades-old algorithms. Thus, there is simply no level of performance that would suggest the utility of a proposed algorithm.

Similarly, if we were told that algorithm *C* was compared to algorithm *D* on these datasets, and algorithm *C* emerged as being an average of 20% better, could we now assume that algorithm *C* really is a better algorithm in general? Again, given what we know about these datasets, even a claimed 20% improvement (larger than the typically claimed margin of improvement) would not imbue confidence. Recall just Fig. 8, on that dataset, if algorithm *C* scored a perfect score, relative to the *claimed* labels, we should regard it as a poor algorithm with *low* sensitivity.

## 3 Introducing the UCR Anomaly Archive

Having observed the faults of many existing anomaly detection benchmarks, we have used the lessons learned to create a new benchmark dataset, The UCR Time Series Anomaly Archive [18]. As we explain below, we have endeavored to make our resource free of the issues we have noted, with one exception. A small fraction of our datasets may be solvable with a one-liner. There are two reasons for this. First, we wanted to have a spectrum of problems ranging from easy to very hard. Second, there *are* occasionally real-word anomalies that manifest themselves in a way that is amenable to a one-liner, and their inclusion will allow researchers to make claims about the generality of their ideas. For example, AspenTech, an oil and gas digital historian, encodes missing data as -9999. If the data is ported to another system and normalized, the exact value of -9999 may change, but such a rapid decrease in value should rightly trigger an anomaly. Such dropouts are generally easy to discover with a one-liner.

To prevent the datasets in the archive reflecting the current authors' biases and interests too much, we broadcasted a call for datasets on social media platforms read by data scientists, and we wrote to hundreds of research groups that had published a paper with "anomaly detection" in the title in the last three years. Alas, this did not yield a single contribution. Nevertheless, the datasets span many domains, including medicine, sports, entomology, industry, space science, robotics, etc.

As we discussed in Section 2.3, we believe that the ideal number of anomalies in a test dataset is one. The reader will be curious as to how we ensured this for our datasets. Clearly, we do not have space to explain this for *each* dataset (although the archive does have detailed provenance and metadata for each dataset [23]). Below we show two representative examples to explain how we created single anomaly datasets.

### 3.1 Natural Anomalies Confirmed Out-of-Band

Consider Fig. 11 which shows an example of one of the datasets in our archive. The first 2,500 datapoints (the '2500' in the file's name) are designed to be used as training data, and the anomaly itself is located between datapoints 5,400 and 5,600 (the '5400_5600' in the file's name) indicate the location of the anomaly.

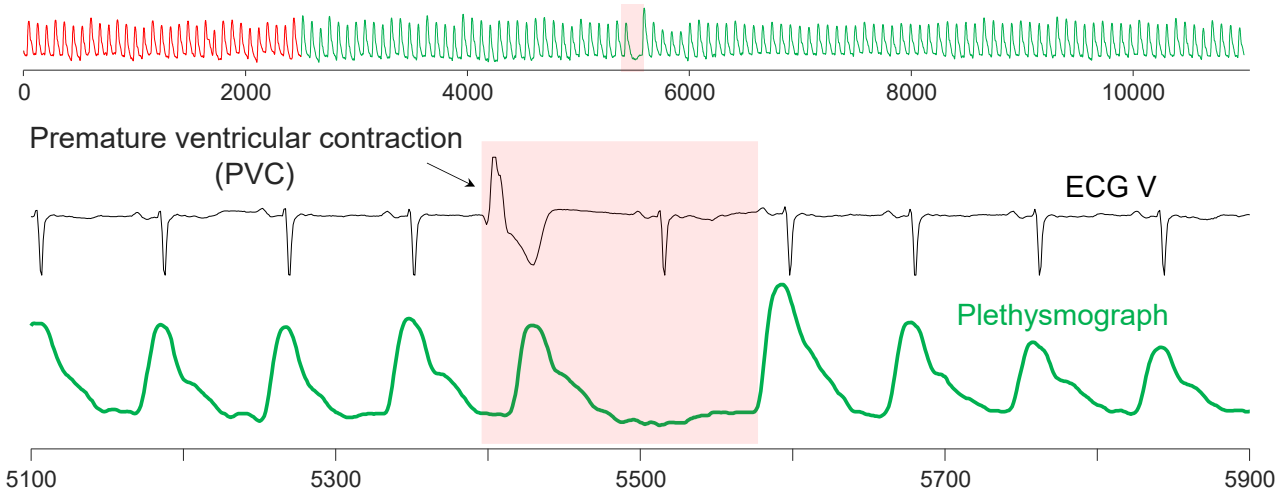

Fig. 11. *(top)* UCR_Anomaly_BIDMC1_2500_5400_5600, a dataset from our archive. *(bottom)* A zoom-in of the region containing the anomaly. A PVC observed in an ECG that was recorded in parallel offers out-of-band evidence that this is a true anomaly.

Here the anomaly is a little subtle. How can we be so confident that is it semantically an anomaly? We can make this assertion because we examined the electrocardiogram that was recorded in parallel. This was the only region that had an abnormal heartbeat, a PVC. Note that there is a slight lag in the timing, as an ECG is an *electrical* signal, and the pleth signal is *mechanical* (pressure). However, the scoring functions typically have a little "play" to avoid the brittleness of requiring spurious precision.

Note that we did not directly create an ECG benchmark here because it is too simple (although we do have a handful of equally simple examples in the archive). We used this general technique, of using obvious out-of-band data to annotate subtle data, to create many of our benchmark datasets.

### 3.2 Synthetic, but Highly Plausible Anomalies

We can also *create* single anomaly datasets in the following way. We find a dataset that is free of anomalies, then insert an anomaly into a random location. However, we want to do this in a way such that the resulting dataset is completely plausible and natural. Fig. 12 shows an example of how we can achieve this.

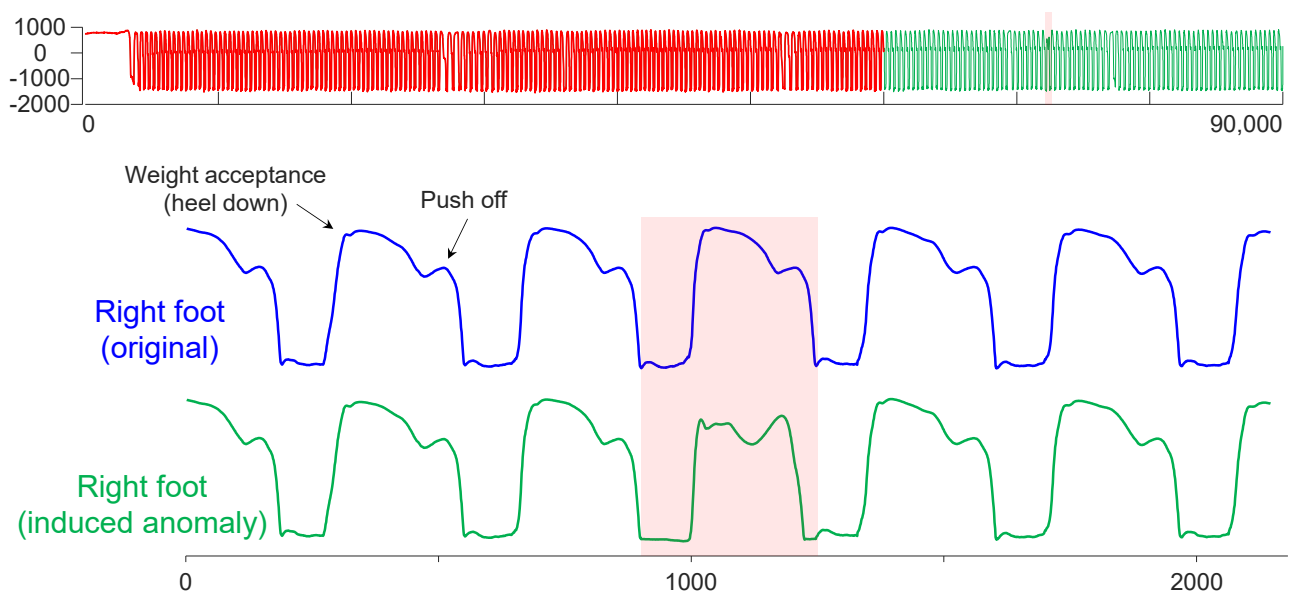

Fig. 12. *(top)* UCR_Anomaly_park3m_60000_72150_72495, a dataset from our Archive. *(bottom)* This individual had a highly asymmetric gait, so we created an anomaly by swapping in a single *left* foot cycle in a time series that otherwise records the *right* foot.

Here we started with a two-dimensional time series, containing the left and right foot telemetry on a force plate. The data came from an individual with an antalgic gait, with a near normal right foot cycle (RFC), but a tentative and weak left foot cycle (LFC). Here we replaced a single, randomly chosen RFC with the corresponding LFC (shifting it by a half cycle length). The resulting dataset looks comply natural, modeling a normal gait, where for one cycle the individual felt a sudden spasm in the leg.

This dataset has another source of viability that happens three or four times. Because the force plate apparatus is of finite length, the gait speed changes as the user circles around at the end of the device. However, we took pains to ensure that both the training and test data have examples of this behavior, so it should not be flagged as an anomaly.

When creating such datasets, we attempted to thread the needle between being too easy, and too difficult. Here we are confident that this example is not impossibly cryptic, as nine out of ten volunteers we asked could identify this anomaly after careful *visual* inspection.

## 4 Recommendations

We conclude with some recommendations for the community.

### 4.1 Existing Datasets should be Abandoned

The community should abandon the Yahoo [5], Numenta [6], NASA [2] and OMNI [3] benchmark datasets. As we have demonstrated, they are irretrievably flawed, and almost certainly impossible to fix, now that we are several years past their creation. Moreover, existing papers that evaluate or compare algorithms primarily or exclusively on these datasets should be discounted (or, ideally *reevaluated* on new challenging datasets).

### 4.2 Algorithms should be Explained with Reference to their Invariances

We would argue that the task of time series *classification* has seen more progress in recent years. In that community, it is understood that it is often useful to discuss novel algorithms in terms of the *invariances* they support [24]. These invariances can include amplitude scaling, offset, occlusion, noise, linear trend, warping, uniform scaling, etc. [24]. This can be a very useful lens for a practitioner to view both domains and algorithms. For example, suppose we wish to classify mosquitoes sex using a Single-Sided Amplitude Spectrum of their flight sounds (as was done in [25]). With a little introspection about entomology and signal processing, we can see that we want any algorithm in this domain to be invariant to the amplitude of the signal. We also want some *limited* warping invariance to compensate for the fact that insect's wingbeat frequency has a dependence of temperature, but not too much warping, which might warp a sluggish female (about 400 Hz) with a much faster male (about 500 Hz). This immediately suggests using a nearest neighbor classifier, with area-under-the-curve constrained DTW (cDTW) as the distance measure. Here, seeing the problem as choosing the right invariances is a very helpful way to both communicate the problem and search the literature for the right solution.

In contrast, one thing that is striking about many recent papers in anomaly detection is that the authors do not clearly communicate under what circumstances the proposed algorithms should work for practitioners that might want to use them. (The work of [26] is a notable exception.) For example, would the ideas in [3] work if my data was similar, but had a wandering baseline that was not relevant to the normal/anomaly distinction?

We suggest that authors could communicate the important invariances with figures.

Consider Fig. 13 *(top)* which shows a one-minute long electrocardiogram that contains a single anomaly (a premature ventricular contraction). The figure also shows the anomaly score from two methods, Telemanom [2] and Discord [19]. Here we are only interested in the relative values, so we omitted the Y-axis, in both cases, the higher values are considered more anomalous. In this example the anomaly is very obvious, and gratifyingly, both methods peak at the location of the anomaly. Visually, we might claim that Discords offer more *discrimination* (informally, the difference between the highest value and the mean values).

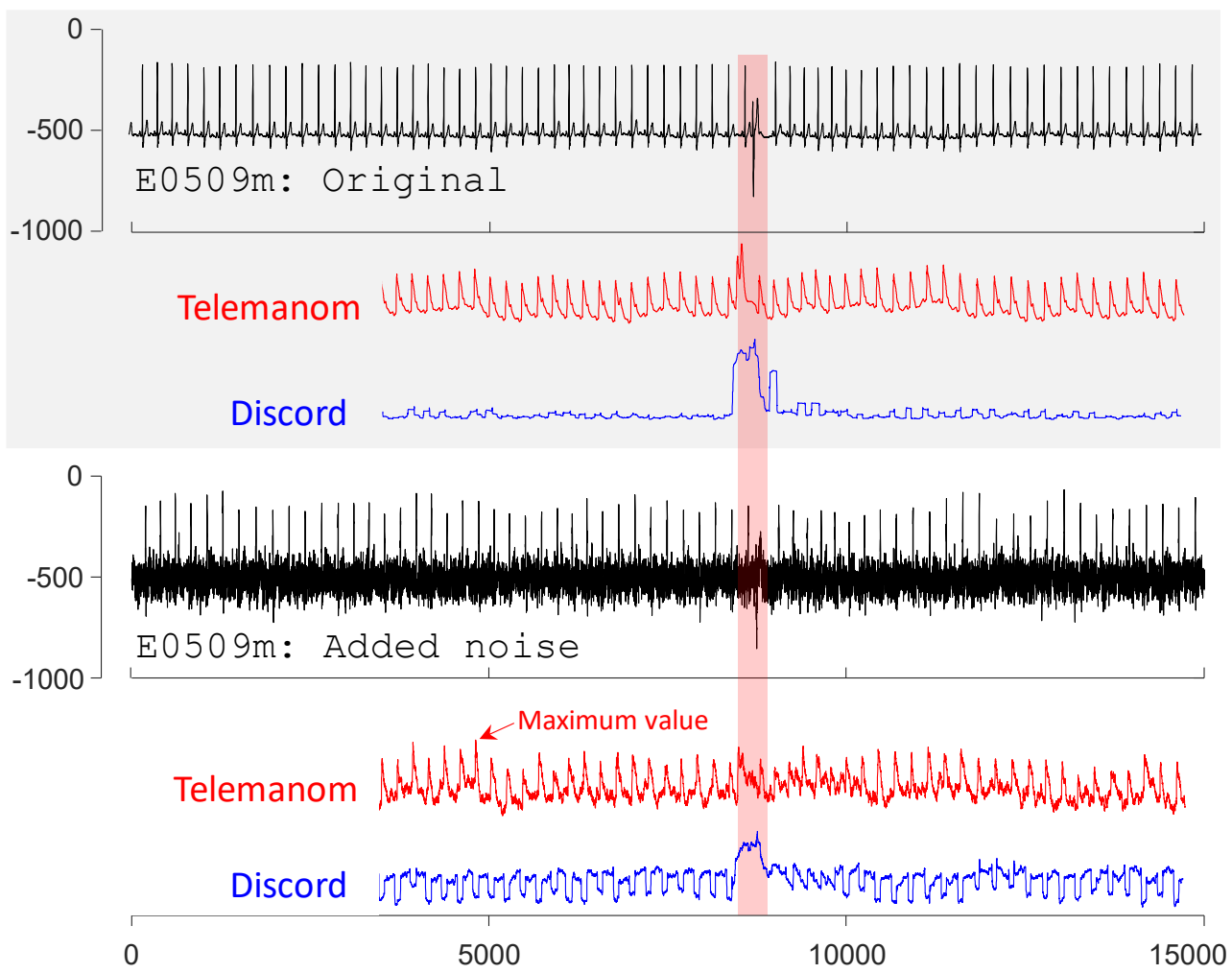

Fig. 13. *(top)* One minute of an electrocardiogram with an obvious anomaly that is correctly identified by two very different methods. Telemanom uses the first 3,000 datapoints from training, using the original authors suggested settings. Discord uses no training data. *(bottom)* The same electrocardiogram with noise added confuses one of the algorithms more that than the other.

In Fig. 13 *(bottom)* we show the same time series, after we added a significant amount of Gaussian noise. The Discord approach now provides less *discrimination*, but still peaks in the right place. In contrast, Telemanom now peaks in the wrong location.

This example suggests that one approach might be better than the other if we expect to encounter noisy data. We are not suggesting that such visualizations *replace* the reporting of metrics such as precision, recall and F1 score, etc. However, for the datasets we consider in this work, those metrics often summarize an algorithm's predictions at just two or three locations. In contrast, the plots shown in Fig. 13 visually summarize the algorithm's predictions at 12,000 locations, and give us a much richer intuition as to the algorithms invariances.

### 4.3 Visualize the Data and Algorithms Output

The point is partly subsumed by the previous point, but worth explicitly stating.

It is very surprising to note that many papers that study time series anomaly detection plot few (as few as *zero*) examples of the time series themselves, in spite of the fact that time series analytics (unlike say protein strings) is inherently a visual domain.

This is more than just a presentation issue; it informs how we should do research. We suspect that some researchers rarely view the time series, they simply pass objects to a black box and look at the F1 scores, etc. One reason we believe this is that the four issues we note in this work are readily visually apparent, they do not need any tools to discover, other than a way to plot the data. For example, the issues with Numenta's NY Taxi dataset discussed in Section 2.4 simply "jump out" of the screen if you plot the data, and the entire data can be comfortably examined on a desktop screen, without even the need for zoom or pan [6]. Yet to our knowledge, no one has noted these problems before.

### 4.4 A Possible Issue with Scoring Functions

In this work we have mostly confined our interest to problems with the current *datasets*. Others have considered problems with current *scoring* functions [20]. However, it would be remiss of us not to note a simple potential issue with scoring functions, especially when comparing rival algorithms. As we noted above, algorithms can place their computed anomaly score at the beginning, the end or the middle of the subsequence. Fig. 13 *(top)* nicely illustrates this. Both approaches can find the obvious anomaly, but Telemanom places its peak earlier than Discords[3]. It is easy to see that unless we are careful to build some "slop" into what we accept as a correct answer, we run the risk of a systemic bias against an algorithm that simply formats its output differently to its rival. As before, *visualization* of the algorithms, together with *visualization* of the acceptable answer range (the red bar in Fig. 13) would go a long way to boost a reader's confidence that the evaluation is fair.

### 4.5 The "deep learning is the answer" Assumption should be Revisited

Many recent papers seem to pose their research question as: "*It is obvious that deep learning is the answer to anomaly detection, here we research the question of what is the best deep learning variant.*" Of course, it is logically possible that deep learning is competitive for anomaly detection, either in general, or in some well-defined circumstances. However, given our findings above, we are not aware of a *single* paper that presents forceful reproducible evidence that deep learning outperforms much simpler methods. For example, Fig. 13 shows that a decades-old method [21] is at least competitive with a highly cited deep-learning approach on one problem, and Nakamura *et al.* [19] provide similar evidence on several datasets. As always, *absence of evidence is not evidence of absence*. Nevertheless, we urge readers to give full consideration to existing methods, which may be competitive, and which are almost always faster, more intuitive, and much simpler compared to deep learning methods that are often slow to train,

[3] This should not be confused with the claim that Telemanom *discovers* the anomaly earlier, which may or may not be true. This is only a minor claim about *formatting* of a particular implementation's output.

opaque and heavily parameter-laden.

## 5 Conclusions

We have shown that the most commonly used benchmarks for anomaly detection have flaws that make them unsuitable for evaluating or comparing anomaly detection algorithms. On a more positive note, we have introduced a new set of benchmark datasets that is largely free of the current benchmark's flaws [23]. However, we do not regard this work as the last word on the matter. Ideally, a committee or a workshop at a conference should gather many diverse viewpoints on these issues, and draft recommendations for the creation of a crowdsourced set of benchmark datasets. We hope this paper will go some way to prod the community into action.

## Acknowledgment

The authors wish to thank all the donors of the original datasets, and everyone that provided feedback on this work. We also wish to thank all that offered comments on an early draft of this work, including Matt P. Dziubinski.

**Renjie Wu** is currently a PhD candidate in Computer Science at the University of California, Riverside. He received his B.S. degree in Computer Science and Technology from Harbin Institute of Technology at Weihai in 2017. His research interests include time series analysis, data mining and machine learning.

**Eamonn Keogh** is a professor of Computer Science at the University of California, Riverside. His research interests include time series data mining and computational entomology.